\documentclass[twocolumn]{svjour3}         
\smartqed  
\usepackage{graphicx}

\usepackage{titlesec} 
\titleformat{\paragraph}[runin]
{\bfseries\scshape}{\theparagraph}{1em}{}
\titlespacing{\paragraph}{0em}{1ex}{.5em} 
\usepackage{todonotes}
\usepackage{placeins} 
\usepackage{booktabs}
\usepackage{tabularx}
\usepackage{multicol}
\usepackage{bm}
\usepackage{hyperref}
\usepackage{url}
\usepackage{balance}
\usepackage{enumitem}
\usepackage{color}
\usepackage{bbding}
\usepackage{nccmath}
\usepackage{amsmath,amssymb}
\usepackage{cite}
\usepackage{makecell}
\usepackage{subfig}
\usepackage{epstopdf}
\usepackage{array}
\usepackage{multirow}
\usepackage{amsfonts}
\usepackage{float}
\usepackage{algorithm}
\usepackage{algorithmic}
\usepackage{microtype}
\usepackage[switch]{lineno}

\begin{document}
\sloppy

\title{ACD: Direct Conditional Control for Video Diffusion Models via Attention Supervision}

\author{Weiqi Li \and
        Zehao Zhang \and
        Liang Lin \and
        Guangrun Wang \thanks{Guangrun Wang is the corresponding author. }
}


\institute{Weiqi Li \at
             Sun Yat-sen University, Guangzhou, China \\
              \email{liwq229@mail2.sysu.edu.cn}
           \and
           Zehao Zhang \at
           Yonsei University, Seoul, Korea \\
              \email{taekho@yonsei.ac.kr}
           \and
          Liang Lin \at
           Sun Yat-sen University; Guangdong Key Laboratory of Big Data Analysis and Processing; X-Era AI Lab \\
              \email{linlng@mail.sysu.edu.cn}
            \and
          Guangrun Wang \at
           Sun Yat-sen University; Guangdong Key Laboratory of Big Data Analysis and Processing; X-Era AI Lab\\
              \email{wanggrun@gmail.com}
}

\date{Received: date / Accepted: date}

\maketitle


\def\eg{\emph{e.g.}}
\def\etal{\emph{et al}}

\newcommand{\EQREF}{Eq.~\eqref}
\newcommand{\EQSREF}{Eqs.~\eqref}
\newcommand{\FIGREF}{Fig.~\ref}
\def\proposed{VB} 
\def\fixcolor{black}
\def\hstate{\bm {\tilde s}}
\def\rstate{\bm s}
\def\jstate{\bm s^{jn}}
\def\jpolicy{\overrightarrow{\pi}}
\def\vpref{v_{\text {pref}}}
\def\vector#1{\mbox{\boldmath $#1$}}
\def\sup#1{^{(\rm #1)}}
\def\sub#1{_{\rm #1}}
\def\supi#1{^{(#1)}}
\def\vct#1{\mbox{\boldmath $#1$}}
\def\eg{{\it e.g.}}
\def\cf{{\it c.f.}}
\def\ie{{\it i.e.}}
\def\etal{{\it et al. }}
\def\etc{{\it etc}}
\newcommand{\argmax}{\mathop{\rm argmax}\limits}
\newcommand{\argmin}{\mathop{\rm argmin}\limits}

\def\Rerr{\Delta \bm r}
\def\Terr{\Delta \bm t}
\def\Xerr{\Delta \bm x}
\def\XerrRel{\Delta \bm {\tilde x}}
\def\Xgt{\dot{\bm x}}
\def\Rgt{\dot{R}}
\def\Tgt{\dot{\bm t}}
\def\arraystretchlen{1.0}

\def\cam{c}
\def\image{\mathcal I}
\def\traj{\mathcal X}
\def\btraj{\mathcal {\bm X}}
\def\keypoints{\mathcal P}
\def\states{\mathcal S}
\def\bstates{\mathcal {\bm S}}
\def\state{\bm s}
\def\ped{\bm x}
\def\pedi{\bm p} 
\def\obs{\bm z}

\def\Fi{\bm F_r}
\def\Fp{\bm F_p}
\def\vpref{\bm w}
\def\ENERGY{{\mathcal E}}

\def\DIFF#1{\textcolor{black}{#1}}
\def\DIFFCR#1{\textcolor{black}{#1}}

\begin{abstract}
Controllability is a fundamental requirement in video synthesis, where accurate alignment with conditioning signals is essential. Existing classifier-free guidance methods typically achieve conditioning indirectly by modeling the joint distribution of data and conditions, which often results in limited controllability over the specified conditions. Classifier-based guidance enforces conditions through an external classifier, but the model may exploit this mechanism to raise the classifier score without genuinely satisfying the intended condition, resulting in adversarial artifacts and limited effective controllability. In this paper, we propose \textit{Attention-Conditional Diffusion (ACD)}, a novel framework for direct conditional control in video diffusion models via attention supervision. By aligning the model’s attention maps with external control signals, ACD achieves better controllability. To support this, we introduce a sparse 3D-aware object layout as an efficient conditioning signal, along with a dedicated Layout ControlNet and an automated annotation pipeline for scalable layout integration. Extensive experiments on benchmark video generation datasets demonstrate that ACD delivers superior alignment with conditioning inputs while preserving temporal coherence and visual fidelity, establishing an effective paradigm for conditional video synthesis.

\end{abstract}

\begin{figure*}
  \includegraphics[width=\textwidth]{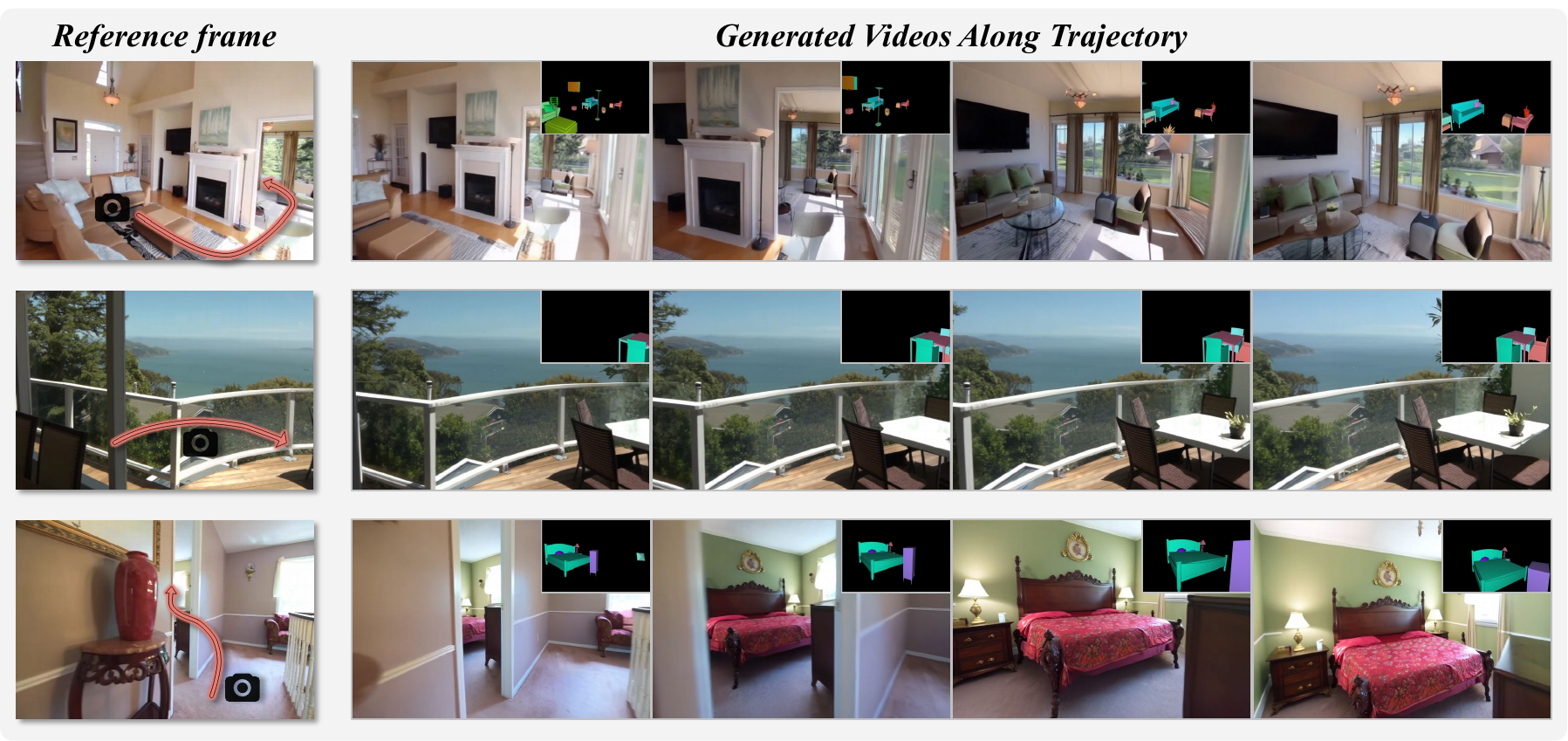}
  \caption{\textbf{Visual results generated by the proposed Attention-Conditional Diffusion (ACD) framework.}
ACD enables direct conditional control in video diffusion models through attention supervision using sparse, 3D-aware object layout signals. Given a single reference image and a sparse object layout with an associated camera trajectory, ACD generates videos that preserve structural semantics and follow the specified camera motion. By applying conditioning at the attention level, the model achieves improved alignment between control inputs and generated content, leading to accurate video synthesis.}
  \label{fig:first_page}
\end{figure*}

\begin{figure}[t]
    \centering
\includegraphics[width=\linewidth]{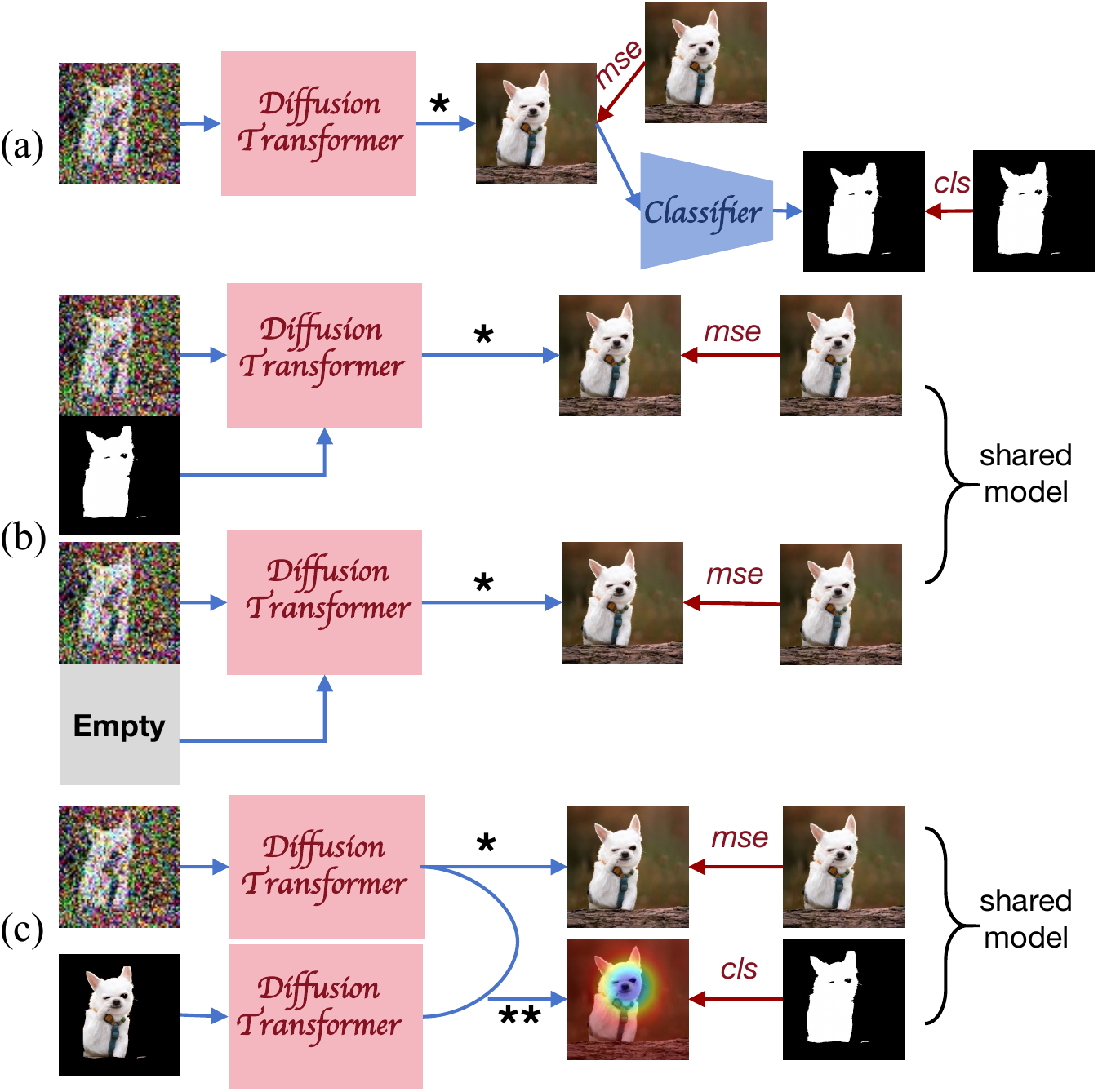}
    \caption{\textbf{Comparison of conditional control strategies in diffusion models.}
(a) \emph{Classifier-based guidance} steers generation using an external classifier, but may produce adversarial artifacts, as the model can increase classifier confidence without genuinely satisfying the intended condition.
(b) \emph{Classifier-free guidance} conditions generation implicitly through joint modeling of data and conditions, yielding strong empirical performance but offering limited fine-grained controllability.
(c) \emph{Attention-conditional control (ours)} applies supervision directly to the model’s attention maps, enabling more direct and semantically grounded alignment between control signals and generated content.
``\(*\)'' denotes the generation of the denoised output (commonly represented as \( x_0 \) or \( x_{\mathrm{start}} \)), either via direct prediction or by estimating the noise and recovering \( x_0 \).
``\(**\)'' indicates the computation of attention between the original latent representation and a masked (or segmented) latent derived from the control signal.
    }
    \label{fig:intro}
\end{figure}

\section{Introduction}

Controllable video generation has emerged as a central goal in generative modeling, driven by the growing demand for precise and semantically grounded control over motion, layout, and scene dynamics in synthesized videos. Recent advances in video diffusion models have enabled impressive visual quality when conditioned on various input signals, such as text prompts, optical flow, or structural layouts. However, achieving \textit{reliable and direct controllability} remains a fundamental challenge. Most existing approaches rely on indirect conditioning mechanisms, which often fail to ensure that the generated video genuinely satisfies the intended conditions.

Early efforts~\cite{chen2024controlavideocontrollabletexttovideodiffusion,zhang2025easycontroladdingefficientflexible,shi2024motioni2vconsistentcontrollableimagetovideo} adapt image-based ControlNet~\cite{zhang2023addingconditionalcontroltexttoimage, mou2023t2iadapterlearningadaptersdig} techniques to the video domain by introducing dense image-level control signals~\cite{koroglu2024onlyflowopticalflowbased,bian2025gsditadvancingvideogeneration}. While these methods provide a degree of controllability, they rely on complex and expensive dense conditioning signals that are difficult to obtain in practice. MotionCtrl~\cite{wang2024motionctrlunifiedflexiblemotion} advances this line of work by using camera poses as explicit control signals, enabling direct manipulation of camera motion. Subsequent methods further refine this idea by incorporating Plücker embeddings of camera trajectories~\cite{he2025cameractrlenablingcameracontrol, bahmaniVD3DTamingLarge2025,bahmaniAC3DAnalyzingImproving2024}. Despite these advances, such approaches often struggle to generalize to complex scenes or long camera trajectories, and their controllability remains limited when scene composition or object-level semantics are involved.

More fundamentally, many controllable video diffusion methods inherit limitations from their underlying guidance strategies. Classifier-free guidance~\cite{hoClassifierFreeDiffusionGuidance2022} enforces conditioning implicitly by learning the joint distribution of data and conditions, which often leads to weak or ambiguous alignment between the generated content and the specified control signals. Classifier-based guidance~\cite{dhariwal2021diffusionmodelsbeatgans, nichol2021improveddenoisingdiffusionprobabilistic}, on the other hand, relies on an external classifier to steer generation. However, this approach is prone to adversarial artifacts: the generative model may exploit the classifier to increase predicted condition likelihood without truly generating content that semantically satisfies the intended condition. As a result, both paradigms struggle to provide effective controllability in practice. A qualitative comparison of these guidance mechanisms is illustrated in Fig.~\ref{fig:intro}.

To overcome these limitations, we propose \textbf{Attention-Conditional Diffusion (ACD)}, a novel framework that enables \emph{direct conditional control} in video diffusion models through attention supervision. Instead of applying conditioning at the output or score level, ACD explicitly aligns the model’s internal attention maps with external control signals, ensuring that conditioning information directly influences the generative process. This design leads to precise and semantically grounded control.

To support ACD, we introduce a \textbf{sparse 3D-aware object layout} as an efficient conditioning representation. Unlike dense image-level controls, this sparse layout naturally captures object geometry and spatial relationships, offering intuitive control over both scene composition and camera viewpoints. We design a dedicated \textbf{Layout ControlNet} to inject this layout information into the diffusion model and develop an \textbf{automated annotation pipeline}~\cite{maninisCADEstateLargescaleCAD2023} to enable scalable data construction. An overview of our framework is shown in Fig.~\ref{fig:first_page}.

Finally, we conduct extensive experiments on benchmark video generation datasets to evaluate controllability, temporal coherence, and visual fidelity. The results demonstrate that ACD consistently achieves superior alignment with conditioning inputs while maintaining high-quality video synthesis, establishing an effective paradigm for conditional video diffusion.

\paragraph{Summary of Contributions.}
Our main contributions are summarized as follows:
\begin{itemize}[leftmargin=*]
    \item We propose \textbf{Attention-Conditional Diffusion (ACD)}, a novel framework for video diffusion that enables \emph{direct conditional control} by supervising the model’s attention maps with external control signals, leading to more reliable semantic alignment between conditions and generated content.
    \item We introduce a \textbf{sparse 3D-aware object layout} as an efficient conditioning representation, enabling intuitive control over object composition and camera viewpoints in video generation.
    \item We design a dedicated \textbf{Layout ControlNet} together with an \textbf{automated annotation pipeline} to facilitate scalable integration of layout-based conditioning into video diffusion models.
    \item We perform extensive experiments on multiple benchmark datasets, demonstrating that ACD consistently improves alignment with conditioning inputs while preserving temporal coherence and visual fidelity.
\end{itemize}

\section{Related Work}

\textbf{Controllable Video Generation. }  
Controllable video generation has seen substantial progress through adaptations of image-based control techniques. Methods like ControlNet~\cite{zhang2023addingconditionalcontroltexttoimage} and T2I-Adapter~\cite{mou2023t2iadapterlearningadaptersdig} introduced external condition encoders to inject structural priors into image diffusion models, enabling fine-grained spatial control. These strategies have been extended to video by leveraging temporal structures, such as optical flow~\cite{koroglu2024onlyflowopticalflowbased}, semantic maps~\cite{chen2024controlavideocontrollabletexttovideodiffusion, zhang2025easycontroladdingefficientflexible}, and keypoints~\cite{huAnimateAnyoneConsistent2024}, to impose constraints across frames. However, such methods typically require dense and carefully aligned condition maps, which limit generalization and scalability. More recent works~\cite{guo2023sparsectrladdingsparsecontrols, Yang_2024, wang2024boximatorgeneratingrichcontrollable} explore lightweight or sketch-based conditioning to lower the barrier for control, but still operate via indirect alignment. Our work differs by introducing \textit{sparse, 3D-aware object layouts} as a minimal yet semantically grounded control interface, and more importantly, enforcing their influence through attention-level supervision rather than output-level conditioning.

\noindent \textbf{Camera Control in Video Diffusion. }  
Control over virtual camera motion is crucial for realistic and expressive video synthesis. Initial attempts~\cite{wang2024motionctrlunifiedflexiblemotion} used simple camera pose sequences to induce basic trajectory control, later extended through more structured representations such as Plücker embeddings~\cite{he2025cameractrlenablingcameracontrol, bahmaniAC3DAnalyzingImproving2024} and 3D-aware embeddings~\cite{bahmaniVD3DTamingLarge2025}. These methods rely on auxiliary control modules inspired by ControlNet to fuse camera cues into diffusion pipelines. While effective in constrained settings, they often struggle to generalize across complex 3D motions or maintain alignment between camera control and scene structure. In contrast, our framework uses sparse 3D object layouts as a unifying representation for both object and camera control, and supervises the model at the attention level to better integrate spatial relationships into the generative process.

\begin{figure*}
    \centering
    \includegraphics[width=\linewidth]{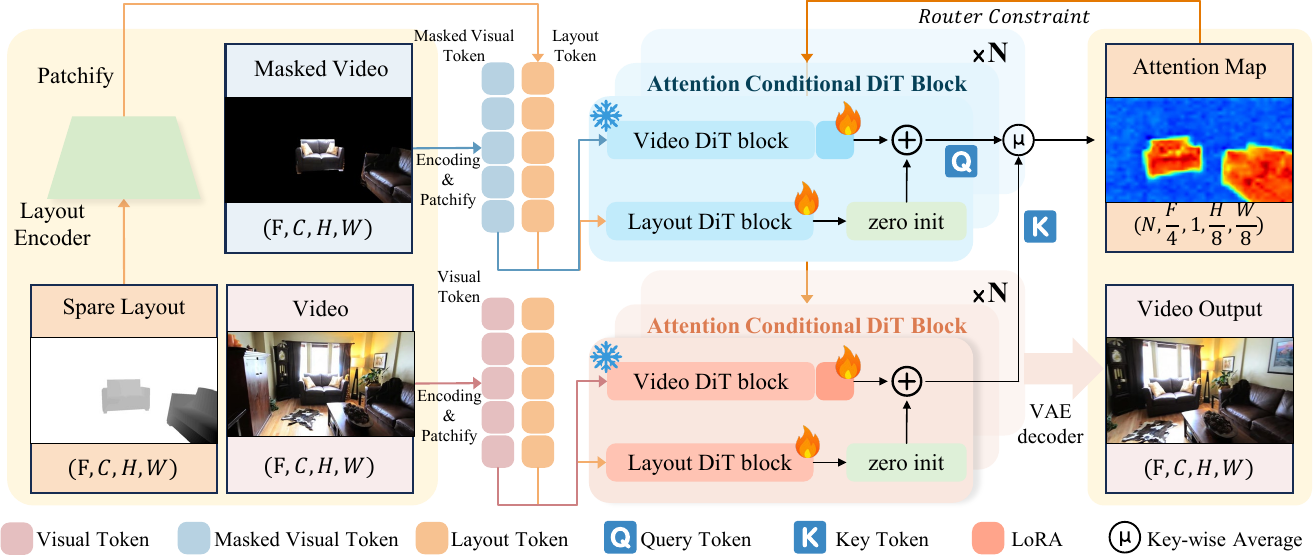}
    \caption{\textbf{Overview of our Attention-Conditional Diffusion (ACD) framework.} The input video and its masked version are encoded into visual tokens, while the sparse 3D-aware object layout is converted into layout tokens. These tokens pass through stacked Attention-Conditional DiT blocks, where a router constraint supervises attention maps between masked and unmasked video tokens. Gradients from this constraint update the model parameters. A VAE decoder then reconstructs the video, enabling ACD to generate outputs that closely follow the given layouts and camera trajectories.}
    \label{fig:framework}
\end{figure*}

\noindent\textbf{Guidance Strategies in Diffusion Models. }  
Guidance mechanisms have been widely adopted to control generation in diffusion models. Classifier-based guidance~\cite{dhariwal2021diffusionmodelsbeatgans, nichol2021improveddenoisingdiffusionprobabilistic} provides explicit direction by training a classifier on noisy samples, but suffers from adversarial behavior. Classifier-free guidance (CFG)~\cite{hoClassifierFreeDiffusionGuidance2022} mitigates these issues by jointly learning conditional and unconditional distributions, enabling smoother interpolation and more practical deployment. CFG has become the standard in both image~\cite{rombach2022highresolutionimagesynthesislatent} and video~\cite{yangCogVideoXTexttoVideoDiffusion2025, podell2023sdxlimprovinglatentdiffusion} diffusion, with many extensions focusing on improved stability, control strength, or multimodal conditioning~\cite{bansal2023universalguidancediffusionmodels}. Despite its effectiveness, CFG only influences model outputs at the sampling level. Our work departs from this paradigm by introducing \textit{attention supervision}, where control signals modulate the internal reasoning of the diffusion model via attention maps, enabling direct alignment between conditions and generated content.

\noindent\textbf{Attention Supervision and Representation-Level Control. }  
Recent interest has emerged in leveraging internal representations—particularly attention maps—for controllable generation~\cite{mou2025dreamo}. These works suggest that modulating attention layers can better guide generation compared to post-hoc or output-level interventions. However, such techniques have largely focused on static image generation or remain auxiliary to broader guidance frameworks. In this work, we formalize attention-level supervision as a core training principle for video diffusion. By jointly optimizing a conditional and an unconditional DiT and enforcing consistency in their attention maps under masked inputs, we achieve \textit{direct conditional control} that propagates semantic intent throughout the generative process.

\section{Method}

Our framework integrates three tightly connected components to enable controllable video generation. We begin by formulating \textbf{3D-Aware Conditional Video Generation} (\S\ref{sec:3D-aware-condition}), which encodes sparse object-centric 3D layouts as explicit control signals, guiding both object placement and camera motion within the generative process. Building on this foundation, we introduce \textbf{Attention-Conditional Diffusion} (\S\ref{sec:ACD}), a mechanism that directly modulates the internal attention maps of a Diffusion Transformer to enforce semantic alignment between these control signals and the synthesized video content. To support training at scale, we develop a \textbf{Dataset Annotation Pipeline} (\S\ref{sec:dataset-pipeline}), an automated system for deriving globally consistent 3D layouts from real-world videos, ensuring that our model is trained on rich, structured supervision.

\begin{figure*}[ht]
    \centering
    \includegraphics[width=\linewidth]{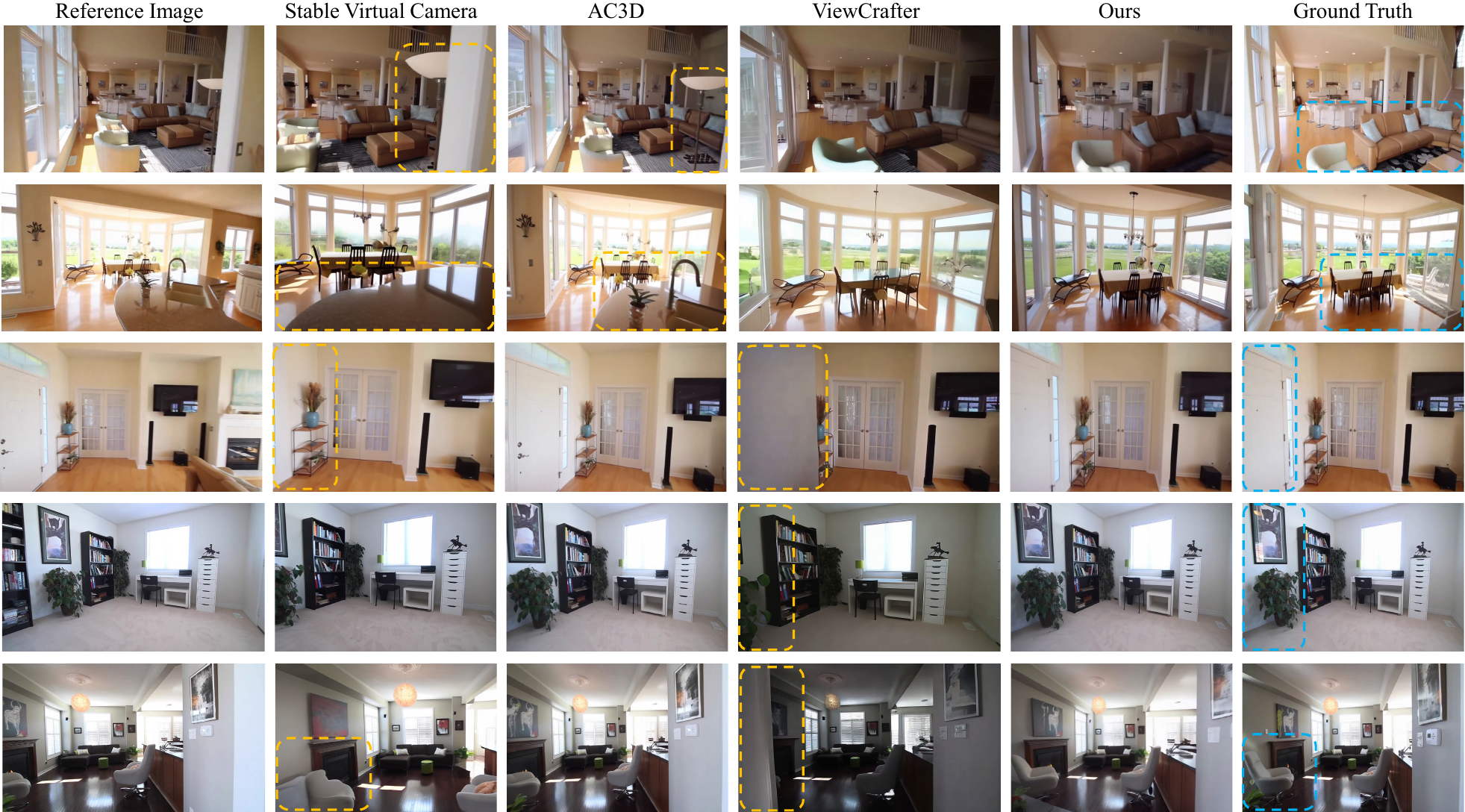}
    \caption{\textbf{Qualitative comparison of video generation results.} Our proposed Attention-Conditional Diffusion (ACD) framework outperforms several state-of-the-art methods, including Stable Virtual Camera, AC3D, and ViewCrafter. The left-most column displays the initial reference images used as input for the generation process. The right-most column shows the ground truth novel views.}
    \label{fig:main}
\end{figure*}

\subsection{3D-Aware Conditional Video Generation}
\label{sec:3D-aware-condition}
The increasing demand for precise control over video content necessitates the development of novel conditioning mechanisms. Our approach introduces a sparse 3D-aware object layout control signal designed to afford fine-grained manipulation of both object and camera trajectories within generated videos. This methodology departs from prior efforts that often rely on dense scene representations, which are computationally intensive and challenging to acquire for generic video data. Instead, we leverage a lightweight yet potent representation of 3D object layouts.

\noindent \textbf{Sparse 3D-Aware Control Signals. } Inspired by CadEstate~\cite{maninisCADEstateLargescaleCAD2023}, we construct a globally consistent 3D object representation from standard RGB videos, circumventing the need for complex, dense scene reconstructions. For annotated objects within a video sequence, we obtain a corresponding CAD model~\cite{chang2015shapenetinformationrich3dmodel} and its precise 3D scene coordinates. This structured object-centric representation offers inherent advantages: it directly encodes 3D spatial information without requiring per-pixel depth maps or intricate mesh reconstructions. From these CAD models, we readily derive two fundamental control signals: sparse depth maps, capturing the spatial extent of objects, and semantic layout information, delineating object identities and their arrangement.

To process these heterogeneous control signals effectively, we employ two distinct lightweight encoders. Each encoder is architected with a series of 3D downsample blocks, which progressively compress the spatial and temporal dimensions of the input. This is followed by a patchification layer that transforms the processed features into a sequence of control tokens, denoted as $c_{layout} \in \mathbb{R}^{N \times d}$, where $N$ represents the sequence length of the tokens and $d$ signifies their embedding dimensionality. These control tokens serve as the compact and informative representation of the desired 3D object layout.

\begin{figure*}
    \centering
    \includegraphics[width=\linewidth]{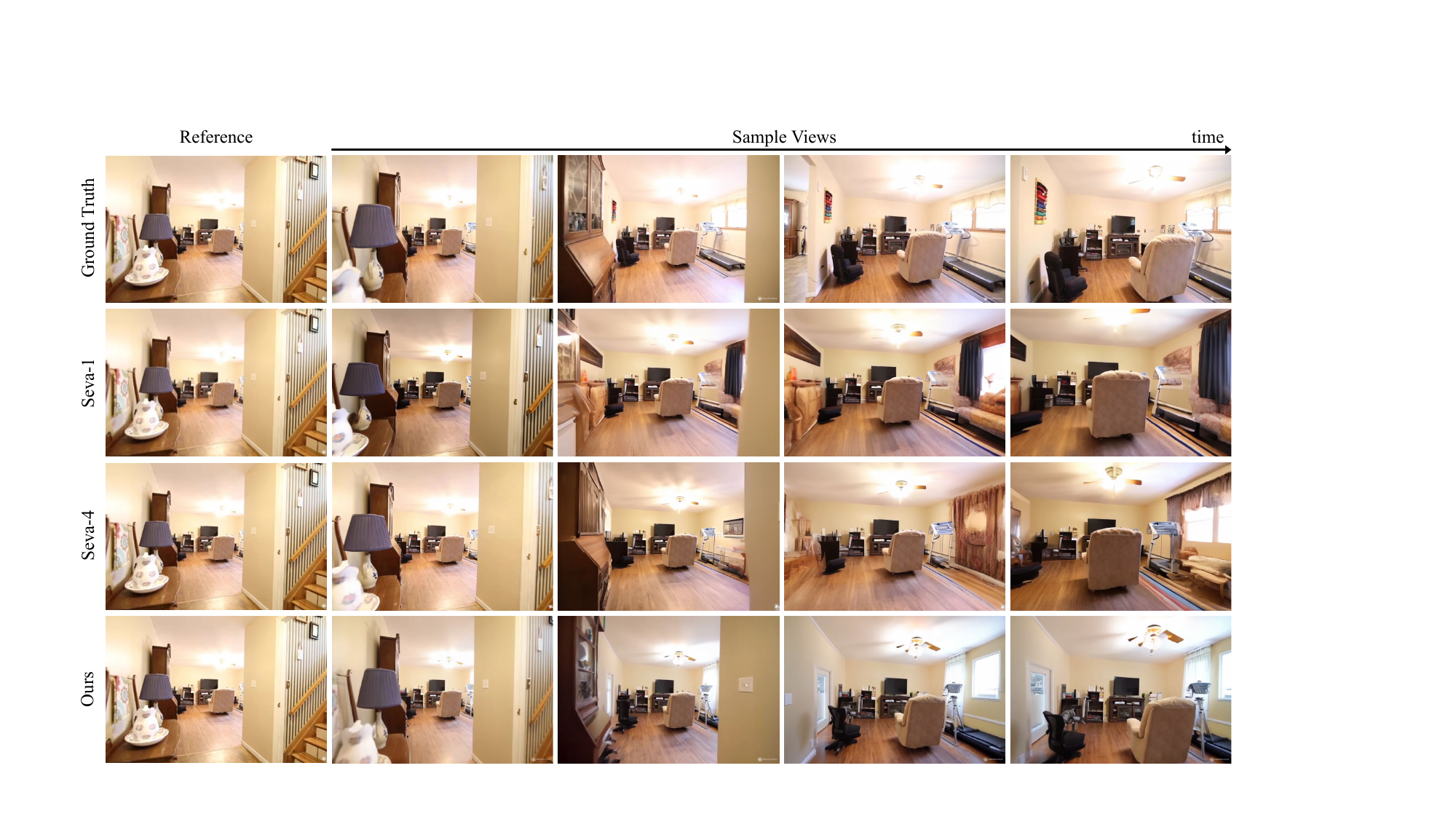}
    \caption{\textbf{Qualitative comparison of video generation results.} Our proposed Attention-Conditional Diffusion (ACD) framework outperforms two variants of Stable Virtual Camera (Seva) on scenes with long camera trajectories. \textit{Seva-1} generates videos conditioned on a single reference image, while \textit{Seva-4} leverages the first four frames as input.}
    \label{fig:more_comp}
\end{figure*}

\noindent \textbf{Foundation Video Model. } Our proposed framework conditions a pre-trained image-to-video (I2V) foundation model on these meticulously designed control signals. The architectural backbone of our I2V model comprises three key components: a 3D causal Variational Auto-Encoder (VAE)~\cite{kingma2022autoencodingvariationalbayes} for learning a latent video representation, a T5 encoder~\cite{raffel2023exploringlimitstransferlearning} for processing textual prompts, and a transformer-based latent diffusion model responsible for the generative process. Within the transformer architecture, each expert block is instantiated as a sequence of 3D full attention mechanisms, expert adaptive layer normalization layers~\cite{yangCogVideoXTexttoVideoDiffusion2025}, and feed-forward networks (FFN). Textual prompts, denoted as $y$, are encoded by the T5 encoder and integrated into the generative process to further guide content synthesis.

The noising and denoising processes within our latent diffusion model follow the principles of Rectified Flow, ensuring a stable and efficient generative trajectory. Given A video latent $z_0 \in \mathbb{R}^{t \times h \times w \times c}$ encoded by 3D Casual VAE, we define a straightforward path between $z_0$ and the noisy latent $z_t$ at a given timestep $t$ as:

\begin{equation}
    z_t = (1-t)z_0 + t\epsilon,
\end{equation}

\noindent where $\epsilon \sim \mathcal{N}(0, \mathbf{I})$ represents standard Gaussian noise. The denoising process is conceptualized as an ordinary differential equation (ODE) that maps the noisy latent $z_t$ back to the clean latent $z_0$:

\begin{equation}
    dz_t = v_{\Theta}(z_t,t,y)dt,
\end{equation}

\noindent where $v_{\Theta}(z_t,t,y)$ represents the velocity field parameterized by the weights $\Theta$ of the denoising network. The training objective is to learn this velocity field through Conditional Flow Matching~\cite{esser2024scalingrectifiedflowtransformers}, which minimizes the discrepancy between the true velocity and the predicted velocity:

\begin{equation}
    \mathcal{L}_{diff} = \mathbb{E}_{t,\epsilon\sim\mathcal{N}(0,\mathbf{I}),z_0}\left[\|(z_0 - \epsilon)-v_{\Theta}(z_t,t,y)\|^2_2\right].
\end{equation}

\noindent This objective effectively trains the denoising network to predict the direction of the flow from noisy to clean data, thereby enabling high-quality video generation.

\noindent \textbf{Sparse Layout ControlNet. } To seamlessly integrate the sparse 3D spatial layouts into the pre-trained I2V model, we introduce a dedicated Sparse Layout ControlNet. This ControlNet is strategically constructed by copying the initial $\mathbf{N}$ DiT blocks (Diffusion Transformer blocks) from the foundational video model. The output of each trainable block $i$ within the ControlNet is first passed through a zero-initialized linear layer, which ensures that the ControlNet does not perturb the pre-trained weights at the start of training. The output of this layer is then element-wise added to the output of the corresponding frozen block in the base I2V model.

\subsection{Attention-Conditional Diffusion}
\label{sec:ACD}

\begin{figure*}
    \centering
    \includegraphics[width=0.85\linewidth]{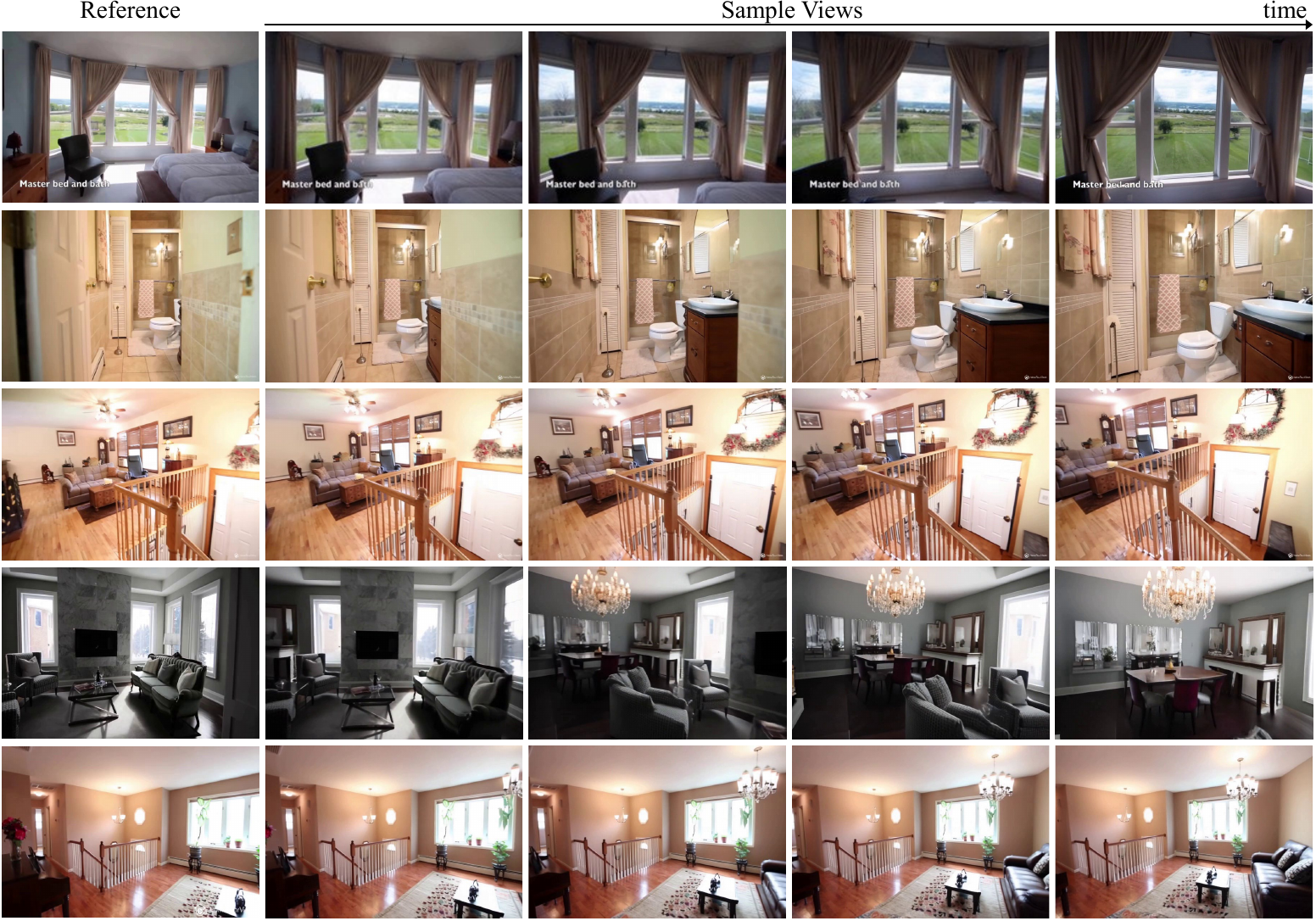}
    \caption{\textbf{Additional Visual Results.} More visual results demonstrating the strong visual quality of our method across diverse indoor scenes.}
    \label{fig:more_visual}
\end{figure*}

\begin{figure*}
    \centering
    \includegraphics[width=0.85\linewidth]{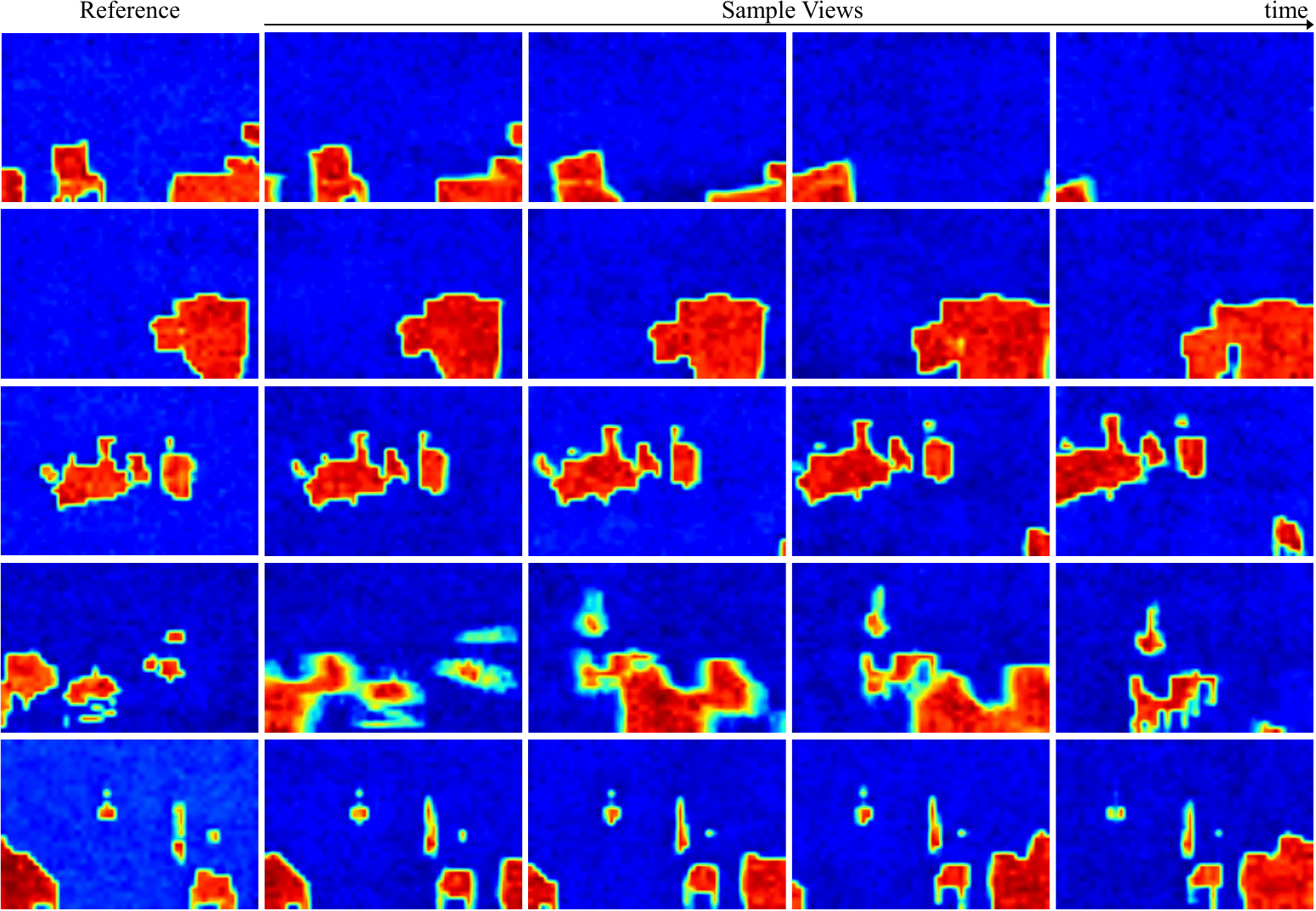}
    \caption{\textbf{Attention Map Visualization.} Corresponding attention maps for the scenes shown in Fig.~\ref{fig:more_visual}, illustrating how our model effectively attends to layout-relevant regions.}
    \label{fig:more_visual_attn}
\end{figure*}

Building upon the advancements in video diffusion models, especially Diffusion Transformers (DiTs), we introduce \textbf{Attention-Conditional Diffusion (ACD)}, a novel framework designed to enhance fine-grained control over video generation by directly modulating the model's internal attention mechanisms, as illustrated in Fig.~\ref{fig:framework}. While ControlNet in Section~\ref{sec:3D-aware-condition} have significantly advanced conditional video generation, they still often struggle with strictly enforcing conditioning semantics. Other guidance methods frequently present limitations in precise control or susceptibility to adversarial artifacts. Our ACD framework addresses these challenges by strategically injecting conditioning signals directly into the attention layers, thereby ensuring a stronger and more reliable alignment between generated content and the intended guidance. Fig.~\ref{fig:more_visual} and Fig.~\ref{fig:more_visual_attn} show the generated video results and the corresponding attention maps results respectively.

\begin{algorithm}[t]
\caption{Attention-Conditional Diffusion}
\label{alg:acd}
\begin{algorithmic}[1]
\REQUIRE Unmasked video $X$, mask $M$ (from sparse 3D layout); \\
Hyperparameters: $\lambda_{\text{diff}}, \lambda_{\text{attn}}$, number of layers $n_{\ell}$, key dimension $d$.

\STATE \textbf{// Encoding and tokenization}
\STATE $Z \leftarrow \text{VAE.encoder}(X)$
\STATE Sample timestep $t \sim \mathcal{U}(\{1,\dots,T\})$, noise $\varepsilon \sim \mathcal{N}(0,I)$
\STATE $Z_t \leftarrow \text{q\_sample}(Z, t, \varepsilon)$ 
\STATE $Z_{\text{mask}} \leftarrow \text{VAE.encoder}(\text{apply\_mask}(X, M))$
\STATE $C \leftarrow \text{LayoutEncoder}(\text{layout})$

\STATE \textbf{// Forward pass (shared DiT)}
\STATE $\mathbf{Q} \leftarrow \text{DiT.forward}(Z_{\text{mask}}, C)$
\STATE $\mathbf{K} \leftarrow \text{DiT.forward}(Z_t, C)$

\STATE \textbf{// Parallel cross-attention loss}
\STATE $\mathbf{M} \leftarrow \text{softmax}\!\left( \frac{\mathbf{Q} \mathbf{K}^\top}{\sqrt{d}} \right)$
\STATE $\mathbf{m}_{\text{resp}} \leftarrow \text{mean}_{\text{query}}(\mathbf{M})$
\STATE $L_{\text{attn}} \leftarrow \text{MSE}\big(\mathbf{m}_{\text{resp}},\ \text{downsample\_and\_pool}(M)\big)$

\STATE \textbf{// Final objective and update}
\STATE $L \leftarrow \lambda_{\text{diff}} L_{\text{diff}} + \lambda_{\text{attn}} L_{\text{attn}}$
\STATE $\text{Optimizer.step}(\nabla_\Theta L)$
\end{algorithmic}
\end{algorithm}

The whole training process is illustrated in Algorithm.~\ref{alg:acd}. Our approach involves jointly fine-tuning two Diffusion Transformer models: a conditional DiT that models the joint probability $p(x, c)$ as discussed in Section~\ref{sec:3D-aware-condition}, and a conditional DiT that processes the masked input $x \odot M$. $x$ represents the video content, $c$ is the conditioning signal derived from our sparse 3D-aware object layout control, and $M$ is a binary mask directly derived from this 3D layout signal. We use a single neural network to parameterize both models, where we process two distinct batches for each training step: one batch containing the unmasked input $x$ and another batch containing a masked input $x \odot M$. While training two separate models is feasible, we share parameters to keep the pipeline simple, reduce training cost, and avoid increasing the overall parameter count. This dual-batch setup allows us to leverage the unmasked input for general content generation while simultaneously guiding the generation process through the masked input and an innovative attention-level constraint.

A cornerstone of our ACD framework is the attention map constraint imposed on the internal representations of the Diffusion Transformer. Within the transformer architecture, cross-attention exists between these two batches. We define the tokens of masked image latent as query $\mathbf{Q_{mask}} \in \mathbb{R}^{N \times d}$ and the tokens of original noisy image latent as key $\mathbf{K} \in \mathbb{R}^{N \times d}$, the cross attention map $\mathbf{M}$ is computed as:

\begin{equation}
\mathbf{M}=\frac{\mathbf{Q}_{mask}\mathbf{K}^{T}}{\sqrt{d}}, 
\end{equation}where $N$ is the number of tokens and $d$ is the dimension of the keys. This attention map essentially dictates how different parts of the conditioning signal influence the generation of the video content. We average the cross-attention map $M \in \mathbb{R}^{N \times N}$ along the query dimension to get the response map $M \in \mathbb{R}^{N}$. By applying a constraint on this internal response map, we directly enforce the semantic alignment between our sparse 3D-aware object layout control and the generated scene structure. Specifically, we define an attention loss, $\mathcal{L}_{attn}$, to minimize the discrepancy between the computed attention map $\mathbf{M}$ and a predefined target map $\mathbf{M}_{target}$:

\begin{equation}
    \mathcal{L}_{attn} = \frac{1}{n_l \times N}\sum_{i=0}^{n_l-1}\sum_{j=0}^{N-1}||\mathbf{M}-\mathbf{M}_{target}||_2^2, 
\end{equation}
where $i$ denotes the layer index, $j$ denotes the token index, and $n_l$ is the total number of attention layers in the DiT. The target map $\mathbf{M}_{target}$ is derived from mask $M$ by applying temporal average pooling and nearest neighbor downsampling. This direct manipulation at the attention level significantly enhances controllability without compromising the inherent generative quality of the model. Furthermore, it effectively mitigates common failure modes observed in prior methods that apply conditioning at a broader, output level, leading to more robust and accurate content generation.

The final loss function for optimizing our ACD framework is a weighted sum of the standard diffusion loss $\mathcal{L}_{diff}$ (which ensures high-quality video generation) and our proposed attention loss $\mathcal{L}_{attn}$:

\begin{equation}
    \mathcal{L} = \lambda_{diff}\cdot \mathcal{L}_{diff}+\lambda_{attn}\cdot \mathcal{L}_{attn}. 
\end{equation}

\begin{figure*}
    \centering
    \includegraphics[width=\linewidth]{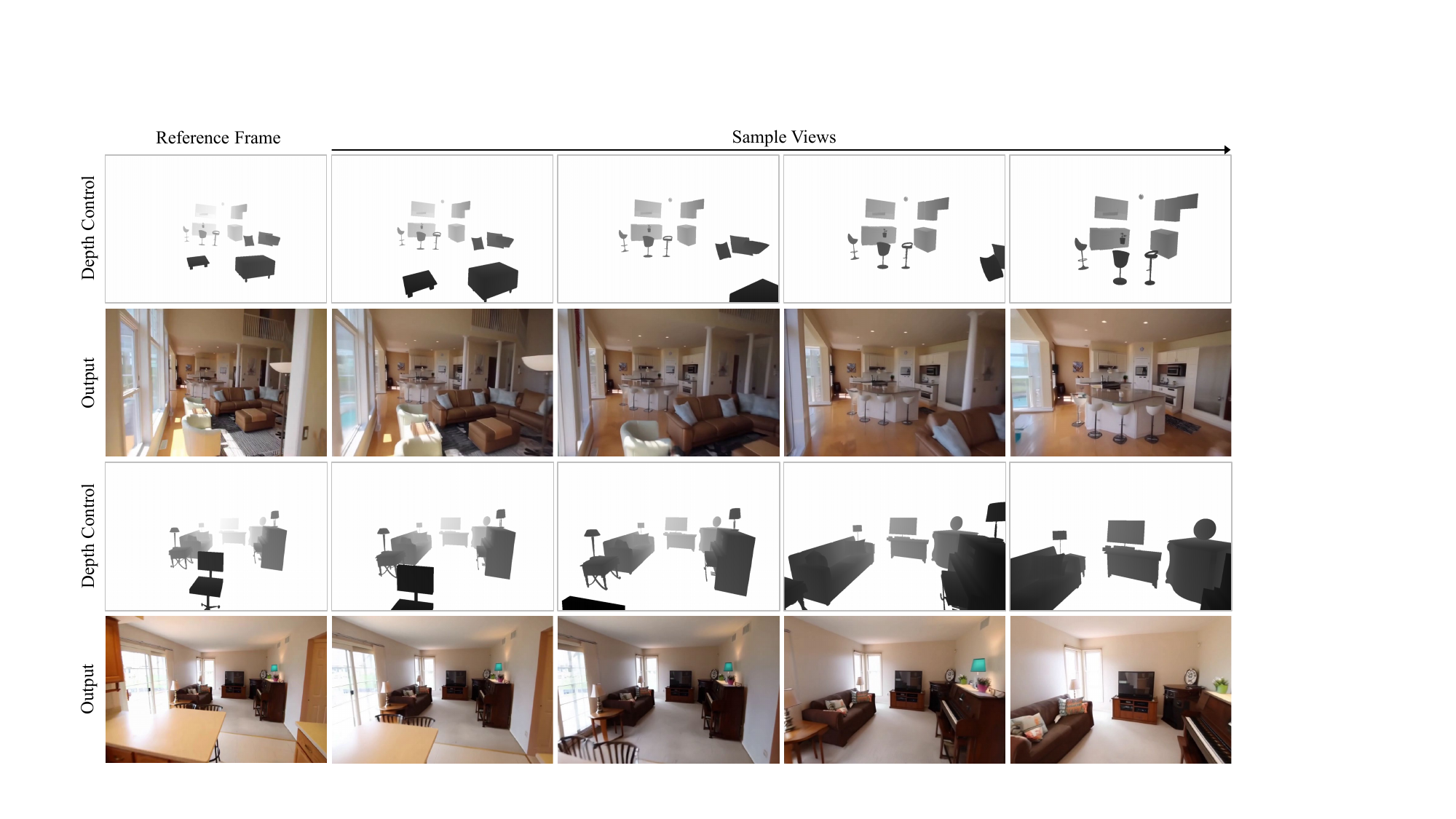}
    \caption{\textbf{Visual Effects of Sparse Depth Control.} Each row shows the sparse depth control signal and the corresponding generated video. }
    \label{fig:depth_control}
\end{figure*}
$\lambda_{diff}$ and $\lambda_{attn}$ are hyperparameters that balance the contribution of each loss component. To efficiently fine-tune the parameters, we employ LoRA (Low-Rank Adaptation)~\cite{hu2021loralowrankadaptationlarge}, a lightweight approach that reduces computational costs when adapting large diffusion models and alleviates overfitting risks on our tailored dataset. This combined loss enables our model to generate high-fidelity videos that are precisely controlled by the sparse 3D-aware object layout, achieving a superior alignment with conditioning inputs while preserving temporal and visual fidelity.

\begin{figure*}
    \centering
    \includegraphics[width=0.95\linewidth]{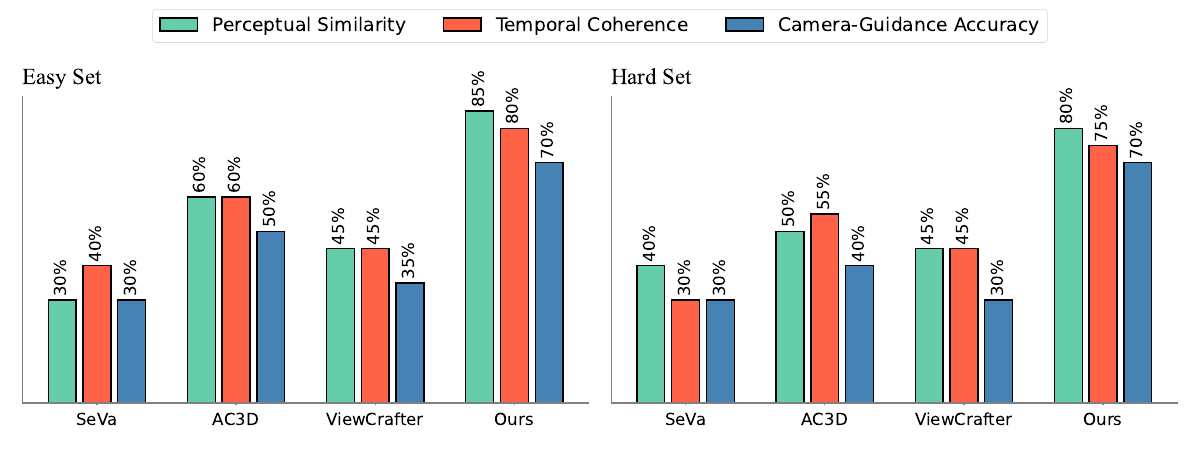}
    \caption{\textbf{User Study Evaluation.} User study ratings for perceptual similarity, temporal coherence, and camera-guidance accuracy on the Easy and Hard evaluation sets. The Easy set represents simpler scenarios with a stride of 2, while the Hard set tests performance under longer temporal ranges and more challenging camera motions with a stride of 5. Our Attention-Conditional Diffusion (ACD) framework consistently received higher ratings than Seva, AC3D, and ViewCrafter across all metrics and both sets.}
    \label{fig:bar_fig}
\end{figure*}

\subsection{Dataset Annotation Pipeline}
\label{sec:dataset-pipeline}

To train our Attention-Conditional Diffusion (ACD) framework, we require a large dataset of videos paired with sparse, 3D-aware object layout annotations. Constructing such data from real-world videos is challenging, as it is difficult to recover reliable 3D information for every object across diverse scenes. To address this, we build on the semi-automatic CADEstate system~\cite{maninisCADEstateLargescaleCAD2023}, which uses a CAD model database to create globally consistent 3D annotations for static indoor videos. We extend this system to handle a larger number of scenes and streamline the annotation process. These annotations are the foundation for generating our sparse depth maps and semantic layout maps, which are fed into the Sparse Layout ControlNet described in Section~\ref{sec:3D-aware-condition}.

\noindent \textbf{Overall Pipeline.} Starting from an indoor video and camera parameters estimated by a Structure-from-Motion (SfM) pipeline~\cite{schoenberger2016sfm}, our annotation system generates for each object a semantic category, a retrieved CAD model, and a full 9-DoF pose (translation, rotation, and independent scaling along three axes). Most steps are automated, while annotators only step in to make targeted corrections, which keeps the process efficient and scalable.

\noindent \textbf{Detection, Tracking, and CAD Retrieval.} The video is first split into shorter clips for easier processing. Objects are detected frame by frame, and detections are linked into consistent tracks over time using multi-object tracking, with occasional human checks to fix broken tracks. For each tracked object, we then retrieve the closest-matching CAD model from ShapeNet~\cite{chang2015shapenetinformationrich3dmodel}, guided by learned shape descriptors and semantic priors. When the system is uncertain, annotators quickly step in to confirm or adjust the choice.

\noindent \textbf{9-DoF Pose Estimation.} After the CAD model is chosen, we refine its placement in the scene by optimizing its 9-DoF pose. This involves minimizing the reprojection error between the CAD model’s silhouette and the object’s appearance in the video frames. Annotators intervene only when the automatic alignment is noticeably off.

\noindent \textbf{Scene Integration and Quality Check.} Once all objects are placed, we merge them into a shared global frame based on the recovered camera trajectory, which resolves scale drifts and ensures all objects sit in a coherent 3D layout. Finally, human reviewers conduct a quick quality check, correcting any remaining inconsistencies. Each review task is small and isolated, so the process can easily be scaled up with crowdsourcing.

\noindent \textbf{Dataset Scope.} Following CADEstate~\cite{maninisCADEstateLargescaleCAD2023}, we annotated videos from RealEstate10K~\cite{zhou2018stereomagnificationlearningview}. Each annotated video provides the exact control signals (sparse depth and semantic layouts) required by our framework, enabling precise and semantically aligned video generation.

\section{Experiments}
\subsection{Experimental Setup}
\textbf{Implementation Details. }
Our framework is built on an image-conditioned, transformer-based video diffusion model, CogVideoX5B-I2V~\cite{yangCogVideoXTexttoVideoDiffusion2025}, which generates videos of 49 frames at a resolution of 480×720. For layout integration, we construct a dedicated Sparse Layout ControlNet by copying 8 DiT blocks from the base model. To implement Attention-Conditional Diffusion (ACD), we design a lightweight LoRA module~\cite{hu2021loralowrankadaptationlarge} with a rank of 128 to modulate the attention layers efficiently without significantly increasing computational cost. We train the model using the Adam~\cite{loshchilov2019decoupledweightdecayregularization} optimizer with a learning rate of 4e-5 on 4 NVIDIA A100 80G GPUs. During inference, we set the classifier-free guidance (CFG)~\cite{chung2024cfgmanifoldconstrainedclassifierfree} scale to 6 and perform sampling with 50 DDIM~\cite{song2022denoisingdiffusionimplicitmodels} steps to balance quality and efficiency.

\textbf{Training Datasets. } To train our framework, we relied on the automated annotation pipeline described in Section~3.3 to assemble a collection of 20K annotated videos. From these videos, we extracted clips using a dynamic sampling stride $s \in \{2, 3, 4, 5\}$, which enabled the dataset to reflect a wide spectrum of temporal scales. This strategy produced segments ranging from roughly 100 to 300 frames.

\begin{table*}[htbp]
  \centering
  
  \resizebox{\textwidth}{!}{
    \begin{tabular}{cccccccccccccccc}
    \toprule
    Dataset & \multicolumn{7}{c}{\textbf{Easy set}}                 &       & \multicolumn{7}{c}{\textbf{Hard set}} \\
\cmidrule{2-8}\cmidrule{10-16}    Method & FID $\downarrow$  & FVD $\downarrow$  & LPIPS $\downarrow$ & PSNR $\uparrow$ & SSIM $\uparrow$ & $R_{err}$ $\downarrow$ & $T_{err}$ $\downarrow$ &       & FID $\downarrow$  & FVD $\downarrow$ & LPIPS $\downarrow$ & PSNR $\uparrow$ & SSIM $\uparrow$ & $R_{err}$ $\downarrow$ & $T_{err}$ $\downarrow$ \\
    \midrule
    ViewCrafter~\cite{yuViewCrafterTamingVideo2024} & 72.745 & 653.91 & 0.430 & 15.97 & 0.592 & 0.092 & 0.179 &       & 88.680 & 796.64 & 0.511 & 13.64 & 0.588 & 0.109 & 0.316 \\
    Seva~\cite{jensenStableVirtualCamera2025}  & 76.328 & 852.79 & 0.422 & 15.88 & 0.594 & 0.106 & 0.256 &       & 85.769 & 1019.21 & 0.492 & 13.35 & 0.549 & 0.144 & 0.392 \\
    AC3D~\cite{bahmaniAC3DAnalyzingImproving2024}  & 64.243 & 786.22 & 0.401 & 16.19 & 0.597 & 0.089 & 0.184 &       & 71.999 & 986.48 & 0.465 & 14.39 & 0.575 & 0.117 & 0.304 \\
    Ours  & \textbf{52.377} & \textbf{377.30} & \textbf{0.285} & \textbf{17.87} & \textbf{0.668} & \textbf{0.078} & \textbf{0.156} &       & \textbf{61.123} & \textbf{481.68} & \textbf{0.341} & \textbf{16.87} & \textbf{0.643} & \textbf{0.094} & \textbf{0.273} \\
    \bottomrule
    \end{tabular}%
    }
  \caption{\textbf{Quantitative Comparison.} Quantitative comparison of ACD with state-of-the-art baselines on the Easy set and Hard set. The Hard set uses a larger stride to test performance under longer temporal ranges and more challenging camera motions.}
  \label{tab:table_main}%

\end{table*}%

\subsection{Comparison with Other Methods}
\textbf{Baselines. } We compare our ACD framework against several state-of-the-art approaches, including ViewCrafter~\cite{yuViewCrafterTamingVideo2024}, AC3D~\cite{bahmaniAC3DAnalyzingImproving2024}, and Seva~\cite{jensenStableVirtualCamera2025}. Among them, AC3D integrates explicit 3D camera control into foundational text-to-video diffusion models, improving spatial consistency and viewpoint manipulation. To enable a direct comparison, we extend AC3D to an image-conditioned video generation setting so it can be evaluated under the same conditions as ACD. For ViewCrafter, we adopt the ViewCrafter25 variant, which is designed to generate sequences of 25 frames, aligning with our evaluation setup. For Seva, we employ the version 1.0 checkpoint, which serves as the officially released model configuration for benchmarking.

\noindent \textbf{Evaluation Metrics. } We assess the quality and controllability of the generated videos through a combination of quantitative metrics. Perceptual similarity between generated content and ground-truth frames is evaluated with PSNR, SSIM~\cite{1284395}, and LPIPS~\cite{zhang2018unreasonableeffectivenessdeepfeatures}. To evaluate visual fidelity and temporal smoothness, we measure the Fréchet Inception Distance (FID)~\cite{heusel2018ganstrainedtimescaleupdate} and Fréchet Video Distance (FVD)~\cite{unterthiner2019accurategenerativemodelsvideo}, which evaluate image realism and temporal consistency across generated sequences. Camera-guidance accuracy is quantified by computing rotation error ($R_{err}$) and translation error ($T_{err}$). We estimate the camera trajectories of generated videos via COLMAP~\cite{schoenberger2016sfm}, normalize all recovered poses to a consistent scale, and compute errors relative to the first frame. Since different baselines produce videos of varying lengths, we report the average metrics across all frames for fair comparison. This combination of metrics provides a balanced view of both the visual realism and the structural faithfulness of our method relative to state-of-the-art baselines.

\begin{figure*}[ht]
    \centering
    \includegraphics[width=\linewidth]{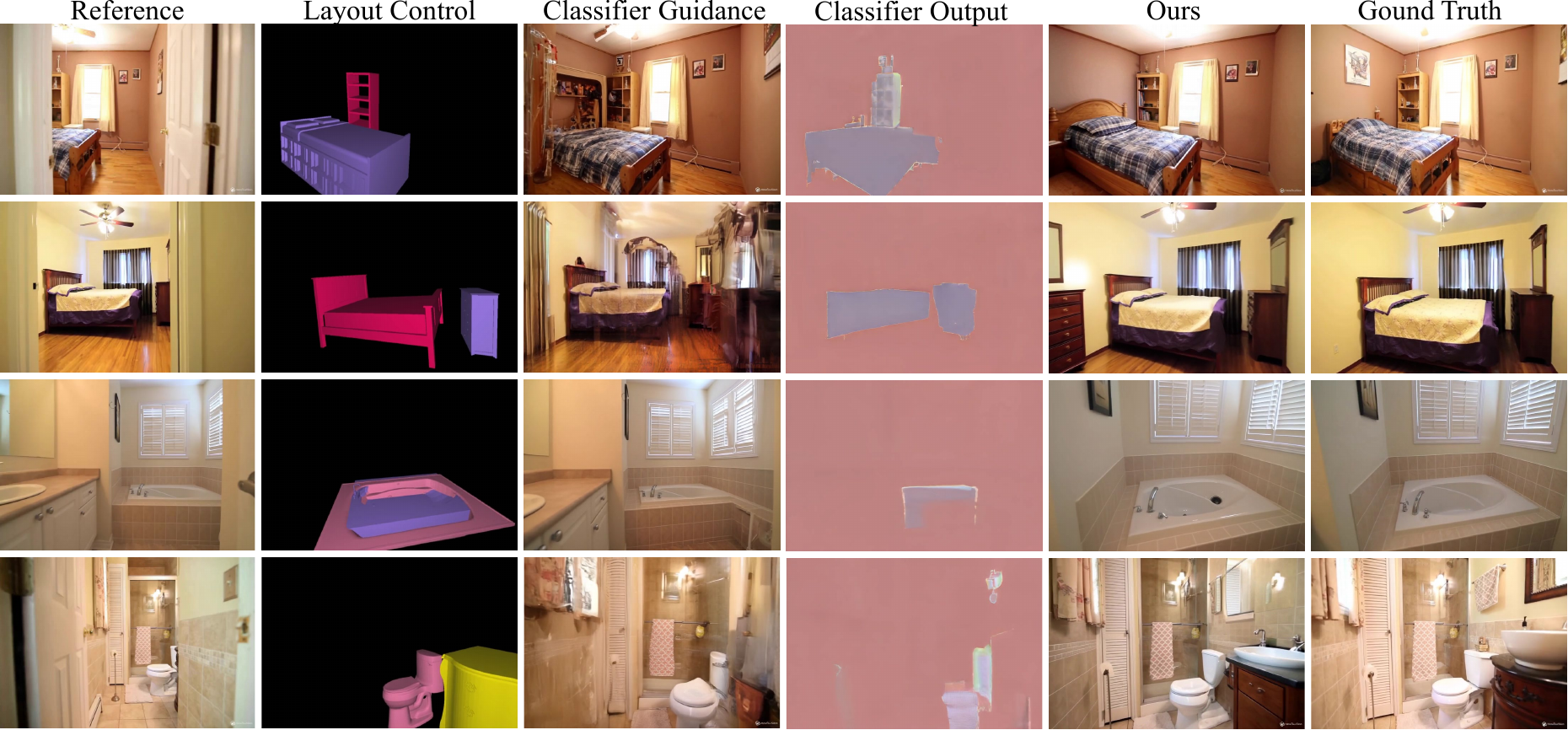}
    \caption{\textbf{Comparison with Classifier Guidance.} ``Classifier Output'' is the depth map predicted by Classifier. Our method avoids adversarial artifacts and better preserves structural alignment with the target layouts.}
    \label{fig:comp_cg}
\end{figure*}

\noindent \textbf{Qualitative Comparisons. }
Fig.~\ref{fig:main} presents side-by-side visual comparisons between our proposed Attention-Conditional Diffusion (ACD) framework and several state-of-the-art baselines. Across a variety of scenes, ACD produces videos that more faithfully preserve the structural semantics of the input reference image and the sparse 3D layout. Unlike AC3D~\cite{bahmaniAC3DAnalyzingImproving2024} and Stable Virtual Camera~\cite{jensenStableVirtualCamera2025}, which frequently exhibit geometric distortions or inconsistent object boundaries when faced with large camera trajectory deviations, ACD maintains coherent spatial relationships and avoids object deformation even under challenging conditions. In addition, ViewCrafter~\cite{yuViewCrafterTamingVideo2024} often introduces unnaturally dark lighting and visible artifacts in certain regions of the frame, further diminishing overall visual quality. Furthermore, the generated camera motion under ACD adheres closely to the prescribed trajectories, yielding smooth, physically plausible transitions without the jitter or drift often observed in competing methods. This observation is further substantiated in scenarios involving extended camera trajectories. As shown in Fig.~\ref{fig:more_comp}, our ACD framework consistently produces sharper, temporally coherent views and maintains coherent structural semantics throughout long-range camera motion. Compared to Stable Virtual Camera methods conditioned on a single reference frame and the first four frames as input (Seva-1 and Seva-4), ACD exhibits greater spatial consistency and fewer geometric distortions, even under limited input supervision.

Visual Effects of Sparse Depth Control. The sparse 3D-aware object layout, used as a control signal, provides an intuitive way to manage camera perspectives. This approach helps generate stable videos, particularly during long-range camera control. As shown in Fig.~\ref{fig:depth_control}, we observe the direct impact of sparse depth control signals on the visual quality of the generated videos. Each row presents a set of sparse depth maps that define the spatial position and size of different objects in the scene, along with the video frames generated by the ACD framework based on these control signals. The sparse information within these depth control maps precisely guides the model to maintain the structural integrity of the object layout in the scene. This method avoids common issues such as geometric distortions or inconsistent object boundaries often found in traditional approaches, especially when dealing with complex or extended camera trajectories.

To further evaluate the effectiveness of our method, we compare it with a Classifier-based Guidance~\cite{dhariwal2021diffusionmodelsbeatgans} implementation. Specifically, we train a video depth estimation model following~\cite{yang2025unifieddensepredictionvideo}, fine-tuned on our constructed layout depth dataset. This model takes video latents as input and predicts the corresponding layout depth. Specifically, we perform channel-wise concatenation of the video latents with depth prediction latents as the input. After training, we backpropagate the mean squared error (MSE) loss through the depth model and obtain the gradients with respect to the video latents, which are then used as guidance signals to adjust the denoising process of video diffusion model. Importantly, the video model employed in this baseline adopts the same backbone architecture as ours for fairness. As illustrated in Fig.~\ref{fig:comp_cg}, the Classifier-based Guidance approach introduces noticeable adversarial artifacts, degrading the visual fidelity of generated videos and weakening control over structural layouts. In contrast, our proposed method achieves cleaner synthesis with stronger alignment to the intended layout signals, highlighting its advantage in balancing controllability and visual quality.

\begin{table}[!t]
  \centering
  
    \begin{tabular}{lrrrrr}
    \toprule
          & \multicolumn{1}{c}{FVD $\downarrow$} & \multicolumn{1}{c}{FID $\downarrow$} & \multicolumn{1}{c}{PSNR$ \uparrow$} & \multicolumn{1}{c}{$R_{err}$$ \downarrow$} & \multicolumn{1}{c}{$T_{err}$ $\downarrow$} \\
    \midrule
    w/o semantic & 510.02 & 69.185 & 16.52 & 0.103 & 0.296 \\
    w/o depth & 513.45 & 72.732 & 15.98 & 0.134 & 0.328 \\
    Ctrl Branch & 480.48 & 65.323 & 16.87 & 0.093 & 0.268 \\
    Post Train & 401.55 & 59.737 & 17.65 & 0.088 & 0.197 \\
    Joint Train & \textbf{357.17} & \textbf{50.248} & \textbf{18.23} & \textbf{0.069} & \textbf{0.134} \\
    \bottomrule
    \end{tabular}%
  \caption{\textbf{Ablation Study.} Ablation study of our proposed ACD framework, comparing different conditioning signals and training strategies.}
  \label{tab:tab_ablation}%
\end{table}%

\noindent \textbf{Quantitative Comparisons. }
We quantitatively evaluate our method against several baselines on two evaluation splits: the Easy set, generated using a stride of 2, and the Hard set, generated using a stride of 5 to capture longer temporal dependencies and more challenging motion scenarios. As summarized in Table~\ref{tab:table_main}, ACD consistently outperforms all competing methods across both sets in terms of visual fidelity, temporal consistency, perceptual similarity, and camera trajectory accuracy. On the Easy set, ACD achieves superior performance across perceptual metrics such as FID, FVD, LPIPS, PSNR, and SSIM, while also demonstrating the most accurate camera rotation and translation estimates. On the more challenging Hard set, ACD maintains its advantage, preserving scene structure, producing temporally coherent frames, and adhering closely to the intended camera motion despite increased scene complexity.

To comprehensively evaluate the perceptual quality and control effectiveness of our proposed Attention-Conditional Diffusion (ACD) framework, we conducted a detailed user study. We carefully selected a total of 20 videos for each method, with 10 videos each from the ``Easy set'' and the ``Hard set'' for evaluation. A panel of 15 participants was asked to evaluate each video. For every video, the participants rated three key aspects: perceptual similarity, temporal coherence, and camera-guidance accuracy. The results of this user study, as visualized in the Fig.~\ref{fig:bar_fig}, demonstrate that our ACD framework consistently received superior ratings across all three metrics and both evaluation sets when compared to competing methods such as Seva, AC3D, and ViewCrafter. On the ``Easy set,'' ACD achieved a dominant lead in perceptual similarity with a rating of $85\%$, significantly surpassing the next-best method, AC3D, which scored $60\%$. This performance gap became even more pronounced on the more challenging ``Hard set.'' ACD maintained a high perceptual similarity rating of $80\%$, a full 30 percentage points higher than AC3D's $50\%$. Most notably, ACD demonstrated exceptional camera-guidance accuracy with a $70\%$ rating, which was more than double the ratings of Seva and ViewCrafter ($30\%$) and well above AC3D's $40\%$. 

\begin{figure*}[ht]
    \centering
    \includegraphics[width=0.95\linewidth]{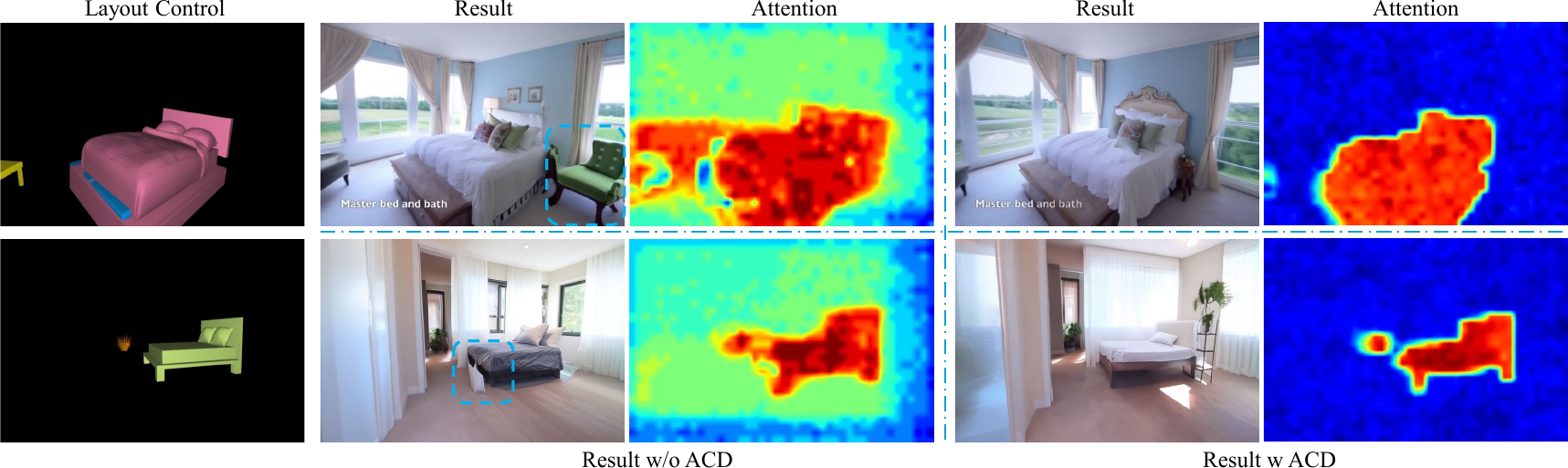}
    \caption{\textbf{Cross-Attention Visualization.} Visualization of cross-attention maps in our Attention-Conditional Diffusion (ACD) framework.}
    \label{fig:ablation}
\end{figure*}

\subsection{Ablation Study}
To thoroughly evaluate the individual contributions of each component within our proposed Attention-Conditional Diffusion (ACD) framework, we conduct a series of ablation studies. These experiments are designed to isolate the impact of key design choices on the model's performance in terms of visual quality, temporal consistency, and adherence to control signals.

\noindent Layout Conditioning Signal. The sparse 3D-aware object layout control signal in ACD is composed of sparse depth maps and semantic layout information. We designed two experiments to assess their individual importance:
\begin{itemize}[leftmargin=*]
\item w/o semantic: This variant represents an experiment where the designed Layout ControlNet does not utilize the sparse semantic map as part of its control signals.
\item w/o depth: In this ablation, the Layout ControlNet was trained without the sparse depth map component of the 3D-aware object layout signal.
\end{itemize}
\noindent Training Strategies. We explored three distinct training paradigms:
\begin{itemize}[leftmargin=*]
    \item Ctrl Branch: This scenario involved using only the ControlNet branch, without the core Attention-Condi-tional Diffusion (ACD) mechanism.
    \item Post Train: In this approach, the ControlNet was first trained independently, and then the ACD mechanism was added and further trained.
    \item Joint Train: This represents our full ACD framework, where the ControlNet and ACD components are jointly fine-tuned from the outset.
\end{itemize}
As shown in Table~\ref{tab:tab_ablation}, removing either of these conditioning signals negatively impacts visual quality and camera accuracy, underscoring the necessity of both semantic and depth information for precise object placement and scene understanding. In addition, our experiments reveal that the Joint Train approach consistently yields the best performance across all metrics. Fig.~\ref{fig:ablation} shows that our method can accurately attend to the provided control signals, leading to improved visual quality in the generated videos. More results are provided on the project website: \url{https://liwq229.github.io/ACD}.

\section{Limitations}
Despite the promising results of our Attention-Conditional Diffusion (ACD) framework, several limitations remain. Our sparse 3D object layout is primarily designed for static indoor scenes, which restricts its applicability to dynamic or outdoor environments where objects may move, deform, or experience significant illumination changes, and extending ACD to such scenarios would require explicitly modeling temporal dynamics and object motion. Moreover, the sparse layout offers only approximate object placement rather than pixel-level alignment, which can lead to misalignments in positioning or scale under complex conditions such as long-range camera trajectories; introducing more accurate geometric priors like dense depth maps or scene flow could help alleviate these issues. Finally, although our automated annotation pipeline reduces the effort of producing sparse layouts, training still depends on annotated layouts, and scaling to larger and more diverse datasets may remain challenging, suggesting that weakly supervised or self-supervised strategies could be explored in future work to mitigate this reliance.

\section{Conclusions}
In this paper, we presented Attention-Conditional Diffusion (ACD), a novel framework for controllable video generation that directly supervises the internal attention maps of diffusion models using sparse 3D-aware layout signals. By enforcing semantic alignment at the attention level, ACD achieves precise control over structural semantics while maintaining temporal coherence and visual fidelity. Extensive experiments demonstrate that ACD consistently outperforms state-of-the-art baselines in both qualitative and quantitative evaluations, confirming its effectiveness and robustness in controllable video synthesis.

\section*{Data availability statement}
The data that support the findings of this study are openly available. This research utilizes the CAD-Estate codebase and dataset schema, which are available at \url{https://github.com/google-research/cad-estate}. Additionally, the 3D object models used for layout construction are sourced from ShapeNet, available at \url{https://shapenet.org/}.

\begin{acknowledgements}
This work was supported in part by the Fundamental Research Funds for Higher Education Institutions allocated to Sun Yat-sen University (Grants 25hytd007 and 2025RGZN009), in part by the Guangdong Provincial High-Level Young Talent Program (Grant 2025HYSPT0707), in part by the Tuoyuan Grant (HT-99982025-0564), in part by the Faculty Start-up Research Fund (Grant 67000-12255002), and in part by the Huawei Strategic Research Institute Talent Fund.
\end{acknowledgements}

\bibliographystyle{spmpsci}      
\bibliography{referencce} 

\end{document}